# Hierarchical Structure Enhances the Convergence and Generalizability of Linear Molecular Representation

Juan-Ni Wu, Tong Wang, Li-Juan Tang, Hai-Long Wu*, Ru-Qin Yu*

*State Key Laboratory of Chemo/Biosensing and Chemometrics, College of Chemistry and Chemical Engineering, Hunan University, Changsha 410082, People's Republic of China*

**ABSTRACT**

Language models demonstrate fundamental abilities in syntax, semantics, and reasoning, though their performance often depends significantly on the inputs they process. This study introduces TSIS (Simplified TSID) and its variants, namely TSISD (TSIS with Depth-First Search), TSISO (TSIS in Order), and TSISR (TSIS in Random), as integral components of the t-SMILES framework. These additions complete the framework's design, providing diverse approaches to molecular representation. Through comprehensive analysis and experiments employing deep generative models, including GPT, diffusion models, and reinforcement learning, the findings reveal that the hierarchical structure of t-SMILES is more straightforward to parse than initially anticipated. Furthermore, t-SMILES consistently outperforms other linear representations such as SMILES, SELFIES, and SAFE, demonstrating superior convergence speed and enhanced generalization capabilities.

# Introduction

Encoding acts as the carrier of information, ensuring that the intended message is accurately transmitted and understood. Effective representation of molecules is a crucial factor affecting the performance of artificial intelligence (AI) models. Unlike natural language processing (NLP) and image recognition, where deep learning has shown exceptional performance, one of the domain-specific challenges for AI assisted molecular discovery is the lack of a naturally applicable, complete and "raw" molecular representation[1].

Molecular descriptors are formal mathematical representations of molecules[2]. Traditionally, molecules are represented as structure diagrams with bonds and atoms. However, molecules are also frequently represented as strings—linear sequences of alphanumeric symbols—known as Linear Molecular Representations (LMRs). Each type of LMRs can be considered a chemical language, as these notations have a defined syntax, meaning not all possible combinations of characters will result in a 'chemically valid' molecule. Additionally, these notations have semantic properties, where the arrangement of elements within the string determines the corresponding molecule's physicochemical and biological properties. This characteristic allows deep learning methods developed for language and sequence modeling to be effectively applied to molecular strings, facilitating chemical language modeling[3][4][5].

Among LMRs, the Simplified Molecular Input Line Entry System (SMILES)[6] is a widely used choice for molecular modeling tasks. However, due to the rule that parentheses and ring identifiers must be paired and the use of the Depth-First Search (DFS) algorithm to parse the molecular graph, SMILES introduces long-range dependencies in grammar, posing challenges for state-of-the-art (SOTA) deep learning models. To address these issues, two alternative atom-based solutions, DeepSMILES (DSMILES)[7], and Self-referencing Embedded Strings (SELFIES)[8], were later proposed. However, both alternatives still struggle with complex grammar, making certain strings more difficult to parse[9][10]

The fragment-based t-SMILES framework, encompassing the TSSA, TSDY, and TSID encodings, was introduced in previous study[11] and has demonstrated significant advantages over classical SMILES, DeepSMILES, and SELFIES. These encodings excel in molecular generation tasks, showcasing superior capabilities in structural representation, novelty, and physicochemical similarity. Such advancements underline the effectiveness of t-SMILES in capturing complex molecular structures and validate its potential as a robust and versatile alternative to traditional molecular representation methods.

SAFE [12] is another fragment-based linear molecular encoding approach that sorts molecular fragments in descending order of atom count, ensuring compatibility with classical SMILES through the extended CXSMILES[13] format. This compatibility enables SAFE to integrate seamlessly with existing SMILES workflows, allowing direct decoding of SAFE strings using the RDKit[14] toolkit when the "allowCXSMILES" parameter is enabled.

In contrast, other fragment ID-based approaches, such as Group-SELFIES[15], encounter some fundamental limitations. These include in-vocabulary (IV) and out-of-vocabulary (OOV) issues, where the encoding struggles to represent fragments that are underrepresented or entirely absent in the training data. Furthermore, these methods face challenges from the curse of dimensionality, stemming from their high-dimensional sparse representations. This sparsity not only reduces model efficiency and accuracy but also hinders generalization capabilities, ultimately limiting the effectiveness of ID-based approaches.

t-SMILES is a tree-based linear molecular description framework that encodes molecules as SMILES-type strings. Within this framework, TSSA, TSDY, and TSID are generated by applying a breadth-first search (BFS) algorithm on a Full Binary Tree (FBT) constructed from a fragmented molecular graph. This study introduces TSIS (Simplified TSID) as a new member of the t-SMILES family, aiming to enhance the framework and facilitate a more comprehensive evaluation using deep generative models (DGMs) such as GPT[16], Diffusion Models (DFM)[17], and Reinforcement Learning (RL) models [18].

A key distinction between TSSA and TSDY/TSID lies in their treatment of fragment connecting points. In TSSA, two fragments share a real atom as the connecting point, while TSDY and TSID utilize dummy atoms (represented by the character "*" with or without an ID) to indicate the bonding between groups. TSIS simplifies the process by directly generating strings through parsing the Acyclic Molecular Tree (AMT) with a BFS algorithm. This is feasible because dummy atoms with IDs (e.g., [n*]) enable the construction of a hierarchical logic that effectively captures the high-level molecular topological structure. Unlike TSID, TSIS eliminates the need for converting AMTs into FBTs, streamlining the encoding process and enhancing encoding and decoding efficiency.

Language is inherently sequential, with word order playing a critical role in the comprehension of meaning[19][20][21][22]. To comprehensively investigate the impact of fragment order on the t-SMILES framework, three encoding algorithms derived from TSIS are introduced: TSISD, TSISO, and TSISR. TSISD employs a DFS algorithm on the AMT. TSISO organizes fragments by length, while TSISR randomizes the fragment order. Syntax and semantics experiments will be conducted using GPT, widely regarded as the leading model in contemporary NLP research.

Systematic experiments reveal, somewhat counterintuitively, that the hierarchical structure of TSID is easier to learn than anticipated. This phenomenon may be explained by the hypothesis that layered architectures reduce the sample complexity of non-trivial learning problems[23]. Furthermore, order-based encodings such as TSISO and SAFE are also learned relatively easily. TSISO demonstrates a simpler learning curve than SAFE due to its reduced reliance on long-term syntactic dependencies. From a semantic perspective, SMILES, TSID, TSIS, and SAFE exhibit comparable performance. However, the SELFIES model generates molecules with fewer aromatic rings, indicating inherent challenges in its encoding logic.

Further analysis using DGMs highlights the superior performance of t-SMILES encodings across multiple dimensions. Specifically, t-SMILES outperforms other representations in exploring the chemical space of the training set, demonstrating higher generalizability in generating novel molecular structures and better alignment with the training data. Additionally, in goal-directed molecular generation tasks, t-SMILES models converge more rapidly than SMILES.

**Methods**

TSIS represents a novel addition to the t-SMILES family, extending its capabilities while adhering to the core principles of the framework. The primary distinction between TSIS and TSID lies in the encoding process: TSIS generates strings by parsing the AMT directly using a BFS algorithm, bypassing the need for conversion to FBT.

Furthermore, three TSIS variants have been developed, each employing a unique method for encoding molecular fragments: TSISD utilizes a DFS algorithm to parse AMT, TSISO orders fragments by length, and TSISR randomly shuffles fragments. These TSIS variants enable a more flexible and nuanced approach to molecular representation, facilitating diverse experimental setups and enhancing the adaptability of t-SMILES for various molecular generation tasks.

The construction of a TSIS string or its variants from a molecule involves two steps:

1. The molecule is fragmented using a selected fragmentation algorithm to create the AMT;

2. The AMT is traversed using the BFS algorithm or other related algorithms to produce the TSIS sequence or its variants.

To reconstruct a molecule from a TSIS and its variants string, follow these two steps:

1. Decompose and reconstruct the AMT from TSIS sequence or its variants.
2. Use the selected algorithm to assemble the molecular fragments, generate the molecular graph, and then optimize it to produce the final molecule.

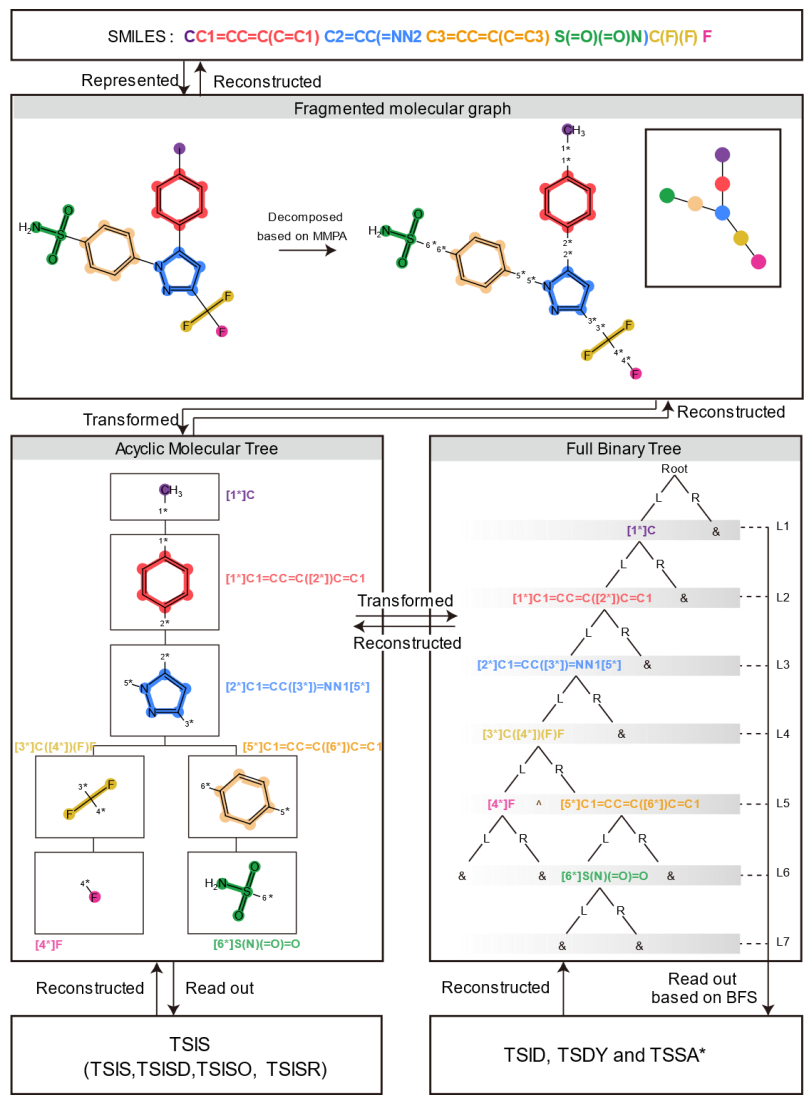

**Fig.1** Overview of the t-SMILES algorithm. In a prior study[11], TSID, TSDY, and TSSA were introduced, deriving from FBT. TSIS and its variants TSISD, TSISO, and TSISR come from AMT. TSIS, TSID,TSDY and TSSA use BFS algorithm to parse AMT or FBT. TSISD use DFS to parse AMT. TSISO sort fragment in order. TSISR shuffles fragments randomly.It should be noted that the logic used to link fragments of TSSA differs from that used for TSDY, TSID, and TSIS.

Table 1 provides a summary of the different t-SMILES codes based on the MMPA fragmentation algorithm applied to the molecule Celecoxib.

Table 1. Codes of TSID and TSIS. TSIS: Breadth-first search algorithm on AMT. TSISD: Depth-first search algorithm on AMT. TSISO: Sorted TSIS based on length of fragments. TSISR: Randomly shuffle fragments of TSIS, the code maybe different based on different random seed.

| Code Algorithm | Celecoxib | Length |
|---|---|---|
| TSID_M (BFS) | [1*]C&[1*]C1=CC=C([2*])C=C1&[2*]C1=CC([3*])=NN1[5*]&[3*]C([4*])(F)F&[4*]F^[5*]C1=CC=C([6*])C=C1&& [6*]S(N)(=O)=O&&& | 114 |
| TSIS_M (BFS) | [1*]C^[1*]C1=CC=C([2*])C=C1^[2*]C1=CC([3*])=NN1[5*]^[3*]C([4*])(F)F^[5*]C1=CC=C([6*])C=C1^[4*]F^[6*]S(N)(=O)=O | 110 |
| TSISD_M (DFS) | [1*]C^[1*]C1=CC=C([2*])C=C1^[2*]C1=CC([3*])=NN1[5*]^[3*]C([4*])(F)F^[4*]F^[5*]C1=CC=C([6*])C=C1^[6*]S(N)(=O)=O | 110 |
| TSISO_M (Order) | [2*]C1=CC([3*])=NN1[5*]^[1*]C1=CC=C([2*])C=C1^[5*]C1=CC=C([6*])C=C1^[3*]C([4*])(F)F^[6*]S(N)(=O)=O^[1*]C^[4*]F | 110 |
| TSISR_M (Random) | [6*]S(N)(=O)=O^[1*]C^[2*]C1=CC([3*])=NN1[5*]^[1*]C1=CC=C([2*])C=C1^[3*]C([4*])(F)F^[5*]C1=CC=C([6*])C=C1^[4*]F | 110 |

## Results

To evaluate the performance of various LMRs, experiments will be conducted on the ChEMBL dataset using GPT, DFM and RL. The evaluation will begin with an analysis of syntax and semantics using GPT models. Subsequently, a comprehensive assessment will be carried out using DGMs across t-SMILES, SMILES, SELFIES and SAFE, focusing on distribution learning metrics, physicochemical properties, and visualized fingerprint space.

**Syntax and semantics**

The true challenge for CLMs lies not in mastering the numerous elementary atom-bonding rules, but in understanding the complex rules that persist even after the basics are acquired[24]. In NLP, syntax and semantics are two fundamental concepts that focus on different aspects of language. **Syntax** refers to the set of rules, principles, and processes that govern the structure of sentences in a language. It concerns how words combine to form phrases, clauses, and sentences, determining the correct word order and hierarchical structure within sentences. **Semantics** deals with the meaning of words, phrases, sentences, and texts. It focuses on how language conveys meaning and how that meaning is interpreted by speakers and listeners.

In SMILES, a syntactic specification[25] outlines how atoms, bonds, parentheses, digits, and other symbols are represented, while a semantic specification describes how these symbols are interpreted to form a valid molecule. For example, the syntax defines how ring closures are written, but the semantics ensures that these closures are correctly paired. Similarly, while the syntax dictates the representation of atoms, the semantics determines whether a ring system is aromatic.

The SMILES specification offers two distinct methods for representing aromaticity: the Kekulé format and the Aromatic Symbols (AS) format. The Kekulé format uses alternating single and double bonds with uppercase symbols for atoms. In contrast, the AS format uses lowercase letters, such as 'c' for aromatic carbon, to represent aromatic atoms without specifying bond types. For example, the Kekulé format of the molecule indane is " C1=CC=CC(CCC2)=C12", while the AS format is " c1ccc2CCCc2c1". Although the Kekulé form obscures the inherent uniformity of bonds in aromatic rings from a chemical perspective[25], it provides a more formatted representation by explicitly detailing partial bond information.

**Syntax**

From a syntactic perspective, t-SMILES uses a tree-based data structure to represent high-level molecular topology. For instance, the skeleton of TSID_M for the molecule Celecoxib can be represented as: 'A&A&A&A&A^A&&A&&&'. This approach may initially seem more complex than SMILES. Additionally, in TSID and TSIS, dummy atoms with IDs (e.g., [n*]) are used to denote joint points, which should appear in pairs across sub-fragments. Moreover, the hidden pattern in TSISO and SAFE involves the order of fragments.

To reveal the impact of syntactic information, such as tree structure and fragment order, experiments will be conducted on the ChEMBL dataset and a subset of 100,000 molecules using the GPT model. The results of these experiments are summarized in Table 2.

**Table.2** Results for the Distribution-Learning Benchmarks on **ChEMBL**(about 1,540,000 molecules) and a subset of 100,000 molecules using **GPT**. Levenshtein distance (LVSD) is calculated to evaluate distance.

| Model | Valid | Uniq | Nov | KLD | FCD | LVSD=0 Or FBT | LVSD=2 | nFrag=1 |
|---|---|---|---|---|---|---|---|---|
| SMILES[R10] (10*100K) | 0.925 | 0.923 | 0.904 | 0.978 | 0.840 | - | - | - |
| SMILES[R1] (1*1540k) | 0.947 | 0.947 | 0.924 | 0.979 | 0.874 | - | - | - |
| SMILES[R10](10*1540k) | 0.980 | 0.979 | 0.907 | 0.992 | 0.906 | - | - | - |
| Kekule[R10] (10*100K) | 0.962 | 0.962 | 0.943 | 0.974 | 0.855 | - | - | - |
| Kekule[R1] (1*1540k) | 0.969 | 0.968 | 0.947 | 0.976 | 0.859 | - | - | - |
| Kekule[R10](10*1540k) | 0.986 | 0.984 | 0.916 | 0.993 | 0.908 | - | - | - |
| SELFIES[R10] (10*100K) | 1.000 | 1.000 | 0.988 | 0.921 | 0.509 | - | - | - |
| SELFIES[R1] (1*1540k) | 1.000 | 1.000 | 0.987 | 0.942 | 0.634 | - | - | - |
| SELFIES[R10](10*1540k) | 1.000 | 1.000 | 0.958 | 0.979 | 0.857 | - | - | - |
| SAFE[R10] (10*100K) | 0.640 | 0.620 | 0.608 | 0.943 | 0.697 | 0.611 | 0.906 | 0.629 |
| SAFE[R1] (1*1540k) | 0.706 | 0.689 | 0.675 | 0.958 | 0.736 | 0.620 | 0.900 | 0.694 |
| SAFE[R10] (10*1540k) | 0.845 | 0.839 | 0.809 | 0.974 | 0.852 | 0.658 | 0.940 | 0.846 |
| TSISO_B[R10] (10*100K) | 1.000 | 0.996 | 0.960 | 0.964 | 0.797 | **0.912** | 0.993 | - |
| TSISO_B[R1] (1*1540k) | 1.000 | 0.997 | 0.963 | 0.966 | 0.834 | 0.910 | 0.992 | - |
| TSISO_B[R10] (10*1540k) | 1.000 | 0.999 | **0.932** | 0.988 | **0.902** | 0.936 | 0.999 | - |
| TSID_B[R10] (10*100K) | 0.999 | 0.998 | 0.972 | 0.971 | 0.832 | 0.987 (128 wrong in 10k) | - | - |
| TSID_B[R1] (1*1540k) | 1.000 | 0.999 | 0.969 | 0.978 | 0.863 | 1.000 (3 wrong in 10k) | - | - |
| TSID_B[R10] (10*1540k) | 1.000 | 0.999 | **0.941** | 0.989 | **0.909** | 1.000 (4 wrong in 10k) | - | - |

Overall, it is somewhat surprising that the FBT structure used by TSID is relatively straightforward to parse, even more so than the order used by TSISO. Analogous to natural language, the symbol "&^" in t-SMILES can be viewed as a marker character for constructing syntax. This finding is consistent with the research in the field of NLP, which suggests that syntactic abilities are acquired with remarkable consistency and reliability early in the pre-training process[26]. Specifically, syntactic rules are often learned within the first 20% of masked language model pre-training, as demonstrated by various syntactic generalization suites[27][28].

Tables 2 illustrates that when the GPT model was trained with 10 rounds on 10,000 training molecules, the FBT was accurately learned and generated with a score of 0.987, indicating a high degree of accuracy. Upon reaching approximately 154,000 training data points, only three generated strings were incorrectly parsed as correct FBT.

A comparison between SAFE and TSISO, both of which sort fragments in order, reveals that TSISO is more efficient at learning syntactic information than SAFE.

Regarding the SAFE algorithm, its valid scores are significantly lower than those of SMILES, indicating that SAFE is also more challenging to parse than SMILES. This observation supports the analysis presented in the discussion section, which suggests that SAFE does not reduce long-term dependency; rather, it may potentially exacerbate it.

Furthermore, when LVSD is set to zero, the resulting lower score of SAFE model indicates that a significant proportion of the generated data is not in the expected order. Additionally, the number of fragments in the generated string (nFrag) is calculated. A value of nFrag greater than one suggests that the SAFE model generates fragmented sets that cannot be parsed as a whole molecule. The lower scores of LVSD and nFrag indicate that the SAFE code is more complex to learn compared to TSISO.

In addition, the classical SMILES format and the Kekule-style approach differ in their descriptions. In general, the Kekule-style approach yields higher scores on validity and the other four metrics. This suggests that the Kekule-style approach may be more efficient than the classical SMILES format in describing molecules in AI generative scenarios.

**Semantics**

Semantic analysis—the study of the meaning of words, phrases, sentences, and texts—is a fundamental aspect of language comprehension. A deep understanding of semantics is essential for reasoning, as it enables the interpretation of context, relationships, and implications within language. For AI models, mastering semantics is crucial because it reflects their ability to grasp and manipulate meaning, which is a key indicator of progress toward human-like intelligence.

In this study, the metric N_AromaticRings (the number of aromatic rings for a molecule), as illustrated in Fig.2, is calculated to evaluate the performance of AI models in handling aromaticity, which is the key semantic concept in LMRs.

With the exception of SELFIES, the remaining five LMRs employ "SMILES" as the fundamental description, demonstrating comparable performance in learning and generating aromatic rings.

The classical SMILES (AS format) and Kekulé format are both relatively straightforward to parse, although the AS format may require slightly more effort at the beginning. However, at the final stage, they achieve almost the same high level of performance.

In terms of SELFIES, which generates more molecules with fewer atomic rings, as illustrated in panel c, where larger values are observed on 0, 1, and 2 and smaller values on 4, 5, and 6, this is consistent with the statement that the approach of SELFIES focus on robustness making some SELFIES strings more challenging to read[10].

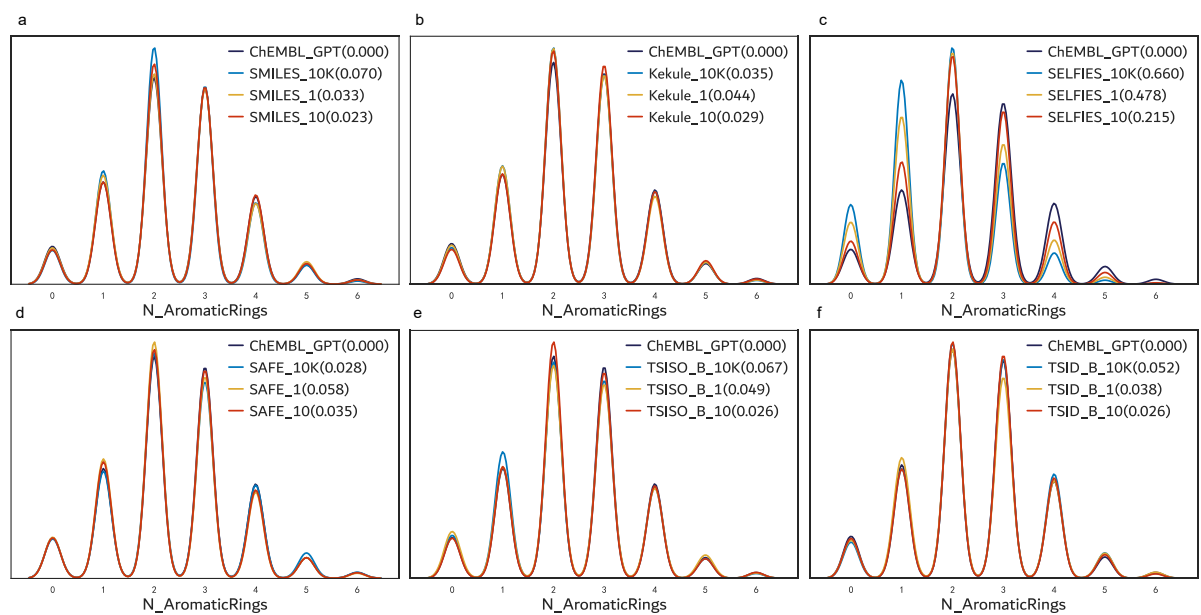

**Fig.2** The number of aromatic rings for a molecule (N-AromaticRings) on **ChEMBL** using **GPT**.

**GPT Models on ChEMBL**

This experiment is designed to conduct a comprehensive and systematic evaluation of the performance of different LMRs using a Transformer-based architectural approach. BRICS, MMPA, and Scaffold are employed as fragmentation algorithms to generate t-SMILES strings. Additionally, the SMILES, SELFIES, and SAFE will serve as baselines for comparison. Given the complexity of SAFE, a larger model ([R15-768]: hidden size of 768, 15 training rounds) is trained to generate more valid molecules. The results are presented in Table 3 and SI.Fig.5.

**Table.3** Results for the Distribution-Learning Benchmarks on **ChEMBL** using **GPT**. R10 means training 10 epochs. 256 or 768 in[R*-*] means the hidden size of model.

| Model | Valid | Uniq | Nov | KLD | FCD | LVSD=0 | LVSD=2 |
|---|---|---|---|---|---|---|---|
| MolGPT[16] | 0.981 | 0.998 | 1.000 | 0.992 | 0.907 | | |
| SMILES[R10-256][11] | 0.980 | 0.979 | **0.907** | 0.992 | **0.906** | - | - |
| DSMILES[R15-256][11] | 0.910 | 0.908 | 0.845 | 0.992 | 0.896 | - | - |
| SELFIES[R10-256][11] | 1.000 | 1.000 | 0.958 | 0.979 | 0.857 | | - |
| SAFE[R10-256] | 0.845 | 0.839 | 0.809 | 0.974 | 0.852 | 0.645 | 0.929 |
| SAFE[R15-768] | 0.911 | 0.906 | 0.746 | 0.992 | 0.896 | 0.661 | 0.934 |
| TSSA_B_[R20][11] | 1.000 | 0.995 | 0.956 | 0.972 | 0.708 | - | - |
| TSDY_B_[R15][11] | 1.000 | 0.999 | 0.960 | 0.977 | 0.854 | - | - |
| TSID_B[R10-256][11] | 1.000 | 0.999 | **0.941** | 0.989 | **0.909** | - | - |
| TSIS_B[R10-256] | 1.000 | 0.990 | 0.932 | 0.990 | 0.902 | - | - |
| TSISO_B[R10-256] | 1.000 | 0.999 | **0.932** | 0.988 | **0.902** | 0.946 | 0.996 |
| TSISR_B[R10-256] | 1.000 | 0.999 | 0.952 | 0.979 | 0.881 | - | - |
| TSSA_M_[R50][11] | 1.000 | 0.996 | 0.970 | 0.982 | 0.808 | - | - |
| TSDY_M_[R15][11] | 1.000 | 0.998 | 0.970 | 0.960 | 0.852 | - | - |
| TSID_M[R10-256][11] | 0.999 | 0.998 | **0.942** | 0.968 | **0.892** | - | - |
| TSIS_M[R10-256] | 1.000 | 0.998 | 0.936 | 0.990 | 0.907 | - | - |
| TSISO_M[R10-256] | 1.000 | 0.998 | **0.938** | 0.991 | **0.907** | 0.938 | 0.995 |
| TSISR_M[R10-256] | 1.000 | 0.996 | 0.970 | 0.969 | 0.842 | - | - |
| TSISD_M[R10-256] | 1.000 | 0.998 | 0.935 | 0.991 | 0.904 | - | - |
| TSSA_S_[R50][11] | 1.000 | 0.998 | 0.977 | 0.966 | 0.795 | - | - |
| TSDY_S_[R15][11] | 1.000 | 0.999 | 0.955 | 0.982 | 0.878 | - | - |
| TSID_S[R10-256][11] | 1.000 | 0.999 | **0.933** | 0.991 | **0.909** | - | - |
| TSIS_S[R10-256] | 1.000 | 0.997 | 0.930 | 0.991 | 0.906 | - | - |
| TSISO_S[R10-256] | 1.000 | 0.997 | **0.940** | 0.991 | **0.907** | 0.985 | 1.000 |
| TSISR_S[R10-256] | 1.000 | 0.998 | 0.955 | 0.984 | 0.891 | - | - |

Table 3 indicates that the model SAFE[R10-256] scores the lowest across all tested models, indicating that the SAFE string is more challenging to parse compared to other LMRs. This supports the findings from the Syntax section, where the logic of SAFE made the analysis of paired numbers and syntax more difficult.

A comparison between the SAFE and TSISO models reveals that the TSISO models outperform the SAFE models across all calculated distribution learning metrics. Furthermore, the analysis of physicochemical properties, as illustrated in SI.Fig.5, indicates that the SAFE model generates a greater number of molecules with smaller aromatic rings and larger rings, which are distinct from the training dataset.

The analysis of t-SMILES code algorithms, including TSID, TSIS, TSISD, TSIO, and TSISR, shows that these algorithms exhibit comparable performance and significantly outperform SMILES, SELFIES, and SAFE. Despite differences in the order of sub-fragments among TSIS, TSISD, TSIO, and TSISR, the CLMs trained on these algorithms achieve relatively better scores. For a detailed comparative analysis, please refer to the discussion section.

In conclusion, the experiment on ChEMBL indicates that TSIS models typically generated novel molecules at a higher rate than SMILES, SELFIES and SAFE models. Furthermore, models trained on TSISO demonstrated superior performance in terms of FCD scores compared with SELFIES and SAFE. In terms of physicochemical properties, both TSIS and TSID demonstrated superior performance in fitting the training data when compared to SAFE and SELFIES. In addition, the TSIS algorithm is flexible in sorting fragments for different purposes, with all variants achieving performance similar to TSID, except for TSISR, which yielded the lowest FCD score among the TSIS variants.

**Diffusion Models on ChEMBL**

The generation of realistic images is considered a primary objective in evaluating the performance of generative models. Diffusion models have emerged as the new state-of-the-art family of deep generative models. They have broken the long-time dominance of generative adversarial networks (GANs) [29] in the challenging task of image synthesis and have also shown potential in a variety of domains, ranging from computer vision, natural language processing, temporal data modeling ,multi-modal modeling, robust machine learning, to interdisciplinary applications in fields such as medical image reconstruction and computational chemistry[30].

Consequently, a published diffusion model DIFFUMOL[17],which combined a diffusion model with 12-layers Transformer architecture to tokenize SMILES and generate molecules with specified scaffolds and properties, is employed for the assessment of t-SMILES codes and other LMRs in this section. Given that DSMILES exhibits relatively lower performance than SMILES and SELFIES, it will not be included in the this experiment due to the computing resources. The detailed results are summaried in Table 4, Fig.3 and Fig4.

**Table.4** Results for the Distribution-Learning Benchmarks on **ChEMBL** using **Diffusion** Models. Two models are trained. Larger models with 12 layers are trained on GPU: NVIDIA Tesla V100s. Smaller model with 8 layers are trained on GPU: NVIDIA GeForce RTX 3090.

| Model | Valid | Uniq | Nov | KLD | FCD | MOSES Nov |
|---|---|---|---|---|---|---|
| DIFFUMOL[L12][17] | 0.943 | 1.000 | 1.000 | - | - | - |
| DFM_SMILES[L12] | 0.935 | 0.935 | 0.913 | 0.942 | 0.700 | 1.000 |
| DFM_SMILES[L8] | 0.876 | 0.875 | 0.858 | 0.910 | 0.608 | - |
| DFM_SELFIES[L8] | 1.000 | 0.999 | 0.983 | 0.911 | 0.500 | |
| DFM_TSDY_B[L8] | 1.000 | 0.998 | 0.943 | 0.930 | 0.654 | - |
| DFM_TSID_B[L8] | 1.000 | 0.998 | 0.937 | 0.944 | **0.723** | - |
| DFM_TSIS_B[L8] | 1.000 | 0.976 | 0.886 | 0.909 | 0.650 | - |
| DFM_TSISO_B[L8] | 1.000 | 0.986 | 0.882 | 0.906 | 0.605 | |
| DFM_TSISR_B[L8] | 1.000 | 0.977 | 0.878 | 0.897 | 0.547 | - |

A 12 layer model DFM_SMILES[L12] was trained on our environment. As anticipated, it exhibites a relatively lower valid score than the published DIFFUMOL[17] as the trianing data is a little more complex and a smaller batchsize is used. Compared to the published baseline model, in this experiment, a smaller diffusion model, DIF_SMILES[L8], which is an 8-layer architecture, is trained as a baseline model for evaluation. All models based on t-SMILES use the same hyperparameters with DIF_SMILES[L8].

When analyzing the distribution learning metrics, Table 4 reveals that a larger SMILES-based model with 12 layers has the potential to outperform a smaller model with 8 layers. However, the smaller TSID-based model exhibits considerably superior performance across all five metrics in comparison to the larger SMILES-based model. Moreover, the smaller TSDY and TSIS models demonstrate enhanced performance on the Novelty-FCD score relative to the SMILES model.These findings suggest that t-SMILES codes are more efficient than SMILES codes, as even smaller models using t-SMILES can extract more useful information than larger SMILES models.

In additoin, the 8 layers SELFIES model DFM_SMILES[L8] exhibits the lowest FCD score among all models. This result suggests that SELFIES is more challenging to parse than SMILES and t-SMILES, and its performance is less effective in maintaining similarity to the training data.

In the t-SMILES family, all the encoding variants, including TSYD, TSID, TSIS, TSISO, and TSISR, demonstrate performance that aligns with their respective design principles. TSDY demonstrates higher novelty score while TSID acheives higher FCD scores. In addition, TSISR consistently attains the lowest FCD score compared to other TSIS variants. This outcome is a reflection of its design logic, where the randomization of fragment order introduces higher variability in the generated structures. While this randomization can enhance diversity, it compromises the consistency and smoothness of the molecular space, leading to a lower FCD score. This result highlights how the encoding strategy directly influences the performance and characteristics of the generated molecular structures.

In evaluating the **physicochemical properties** in Fig.3, the figures of MolWt and N_Atoms reveal that SMILES, SELFIES and t-SMILES models produce a pronounced Gaussian distribution, which deviates significantly from the training data and the observoaiton of GPT model. This observation aligns with the diffusion model's principle of sampling latent variables from a Gaussian distribution. Moreover, these figures also provide insight into the reason for the lower FCD scores observed.

Although the FCD scores are relatively lower, the SAScore figure indicates that the DFMs of t-SMILES and SMILES achieve comparable positive results. This is not the case with the GPT models of SELFIES and TSSA, which exhibit relatively lower performance in terms of both FCD and SAScore.

However, the SELFIES-based DFM continues to yield molecules with a lower SAScore, accompanied by heightened difficulty. This is analogous to the behavior observed in GPT models. For a detailed comparative analysis, please see the discussion section.

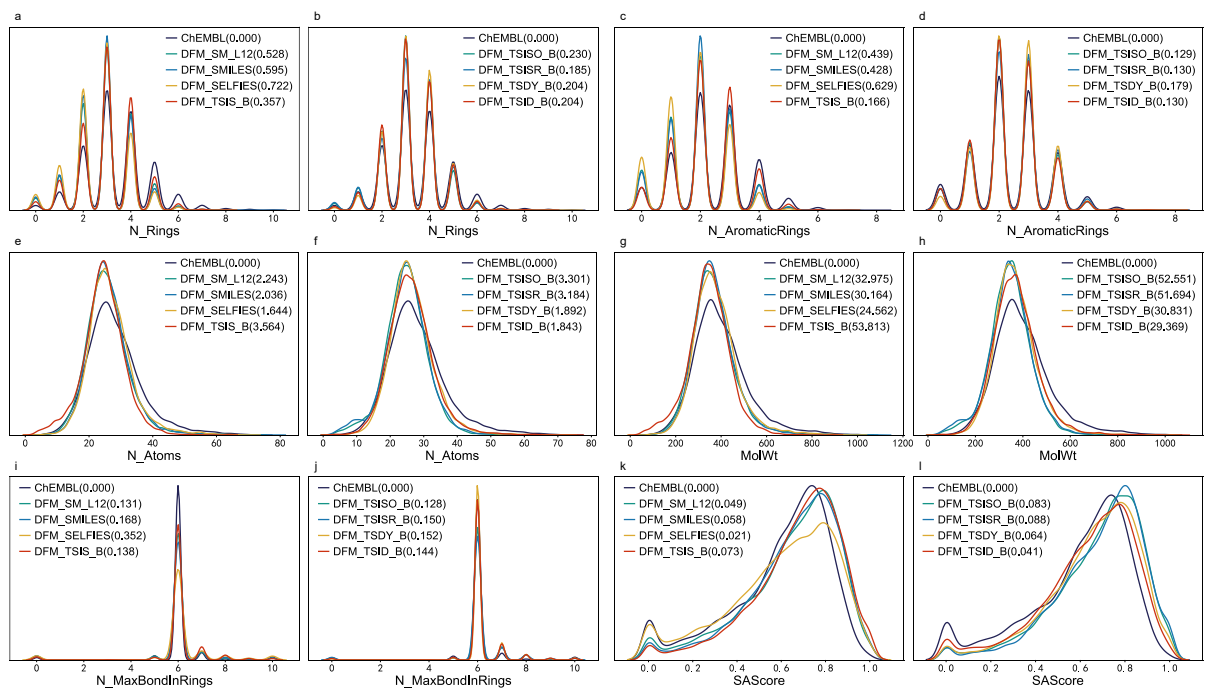

**Fig.3** Physicochemical properties on **ChEMBL** using diffusion model.

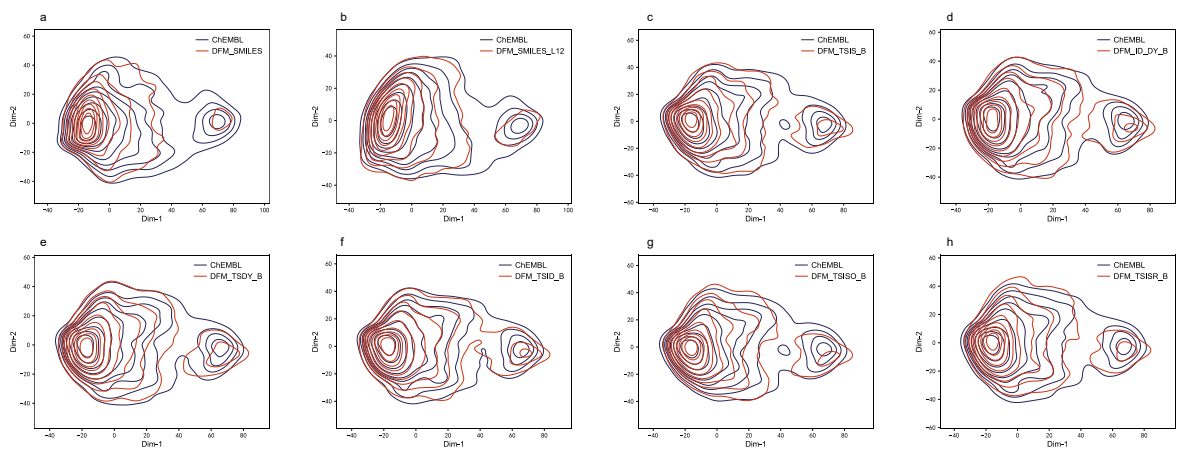

**Fig.4** Visualization of the training data and the generated molecules on ChEMBL using diffusion model for different LMRs. ISOMAP[31] is used as a dimensionality reduction algorithm, RDKit topological fingerprint[14] is used as a fingerprinting algorithm.

Regarding the N-AromaticRings metric, the SMILES and SELFIES models tend to produce a higher number of molecules with fewer aromatic rings. In contrast, the t-SMILES-based model generates a greater number of molecules with a significantly higher count of aromatic rings. Furthermore, it is evident that the SELFIES model produces a greater number of molecules with a lesser 6-bond ring, as illustrated by the N_MaxBondInRings figure.

As shown in Fig.4, the **visualized fingerprint space** reveals significant differences in the exploration of chemical space among the various models.

The SMILES and SELFIES models exhibit a narrow chemical space, reflecting the lowest generalizability, which is consistent with its lower Novelty-FCD scores. In contrast, the TSID-based model explores the widest chemical space, demonstrating the highest generalizability. This is reflected in its superior Novelty-FCD scores, indicating that TSID is more effective at generating diverse and novel molecules that remain consistent with the training data's distribution. The TSDY and TSIS models follow, showing slightly narrower chemical spaces but still outperforming SMILES in terms of diversity and generalization.

**Reinforcement Learning Models for Goal-Directed Molecular Generation**

For goal-directed molecular generation, in addition to the few-shot and transfer learning models demonstrated in the previous studies[11], RL models offer another highly effective approach. As a principle verification, this section uses the RL model published in REINVENT[18] as a baseline to evaluate both SMILES and t-SMILES.

The objective of this experiment is to generate structures similar to the target molecule, Celecoxib. Some key properties of the target molecule include: MolWt (381.379), n_Atoms (26), SA_Score (0.816), and FCSP3(0.118). The properties of the generated molecules are shown in Fig.5,while the generated molecular structures are depicted in Fig.6.

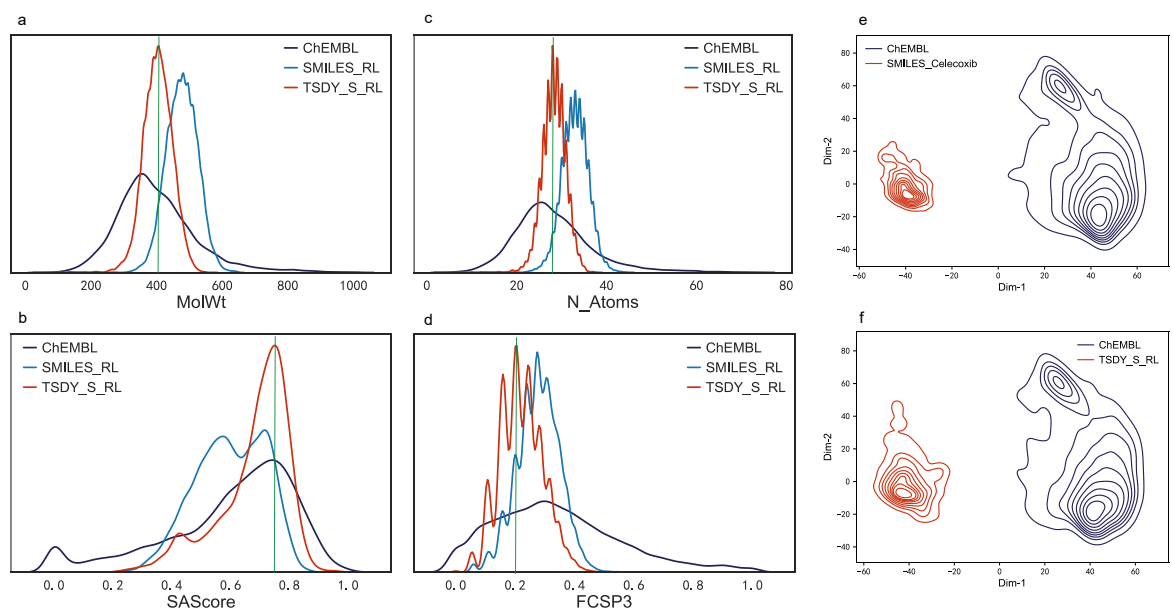

**Fig.5** Physicochemical properties and fingerprint space of similarity guided molecular generation task.

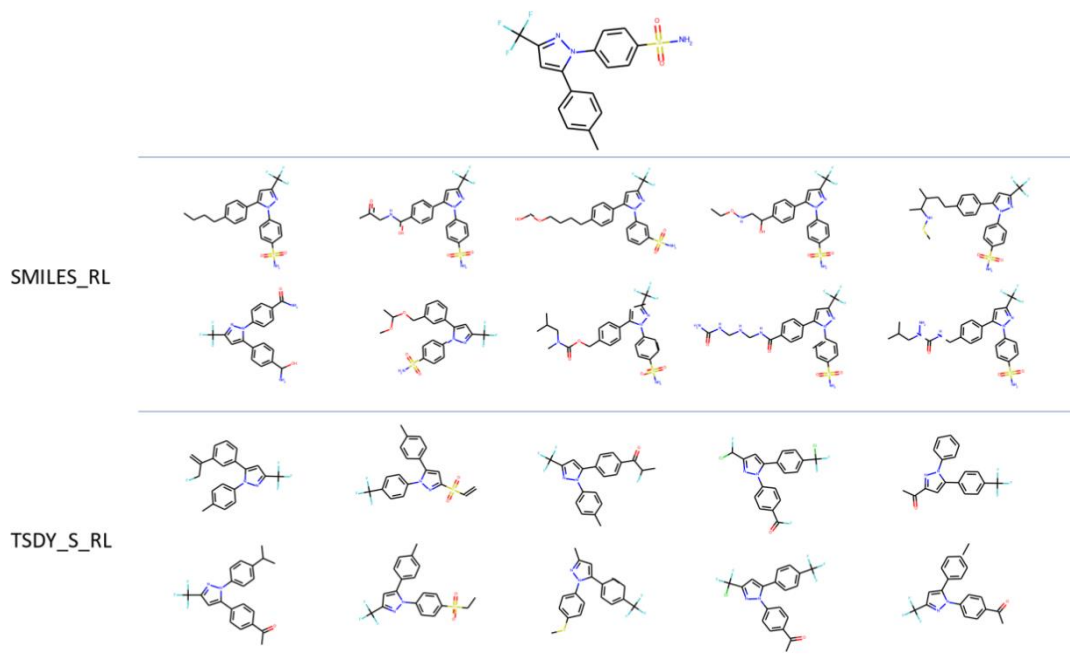

**Fig.6** Molecules generated by similarity guided molecular generation models.

As illustrated in Fig.5, the TSDY_S model demonstrates a more rapid convergence to the physicochemical properties of the target molecule in comparison to the SMILES model after 500 training epochs. This suggests that the TSDY_S model exhibits a superior capacity to capture the molecular structure and physicochemical properties during training, thereby enabling it to generate molecules that align with the target properties with better efficiency.

As illustrated in Fig.5, despite the set of molecules similar to the target molecule lying outside the training data, the models successfully generate a set of molecules that meet the specified criteria. Furthermore, the TSDY_S model produces molecules that exhibit a more extensive chemical distribution compared to the SMILES model, indicating enhanced exploration and generalization capabilities and reflecting a higher degree of molecular diversity.

As illustrated in Fig.6, the similarity-guided reinforcement learning model exhibits the capacity to perform specific skeletal hopping. By defining a well-designed reward function, the model not only retains the primary framework of the molecule but also facilitates the generation of reasonable structural transformations. This approach enables the exploration of diverse chemical spaces while preserving the core structure. Furthermore, the molecules generated by the TSDY_S model exhibit structures that are closer to the target molecule than those generated by the SMILES model, thereby demonstrating superior accuracy and effectiveness in molecular design.

In conclusion, the findings of this experiment indicate that the TSDY_S model exhibits superior convergence and generalization capabilities compared to the SMILES model.

## Discussion

**Distinctive Properties**

The t-SMILES framework possesses some distinctive properties that set it apart from other LMRs. The previous studie[11] has explored its token distribution and nesting depth, highlighting its structural and hierarchical advantages. This section will delve deeper into the t-SMILES framework, exploring its relevance and performance in relation to some of the most significant concepts in deep learning, such as compositional generalization, hierarchical modeling, and representational efficiency.

**Decomposition and Hierarchical learning**

According to cognitive psychology and related disciplines, the development of complex problem-solving behaviour in biological agents depends on hierarchical cognitive mechanisms[32]. One of the core strengths of deep learning lies in its hierarchical structure and layer-wise feature extraction. This capability allows deep learning models to automatically learn and extract high-level abstract features from raw data, which is essential for many tasks that require understanding complex patterns, such as image recognition, natural language processing, and molecular generation. The efficacy of hierarchical learning ability hinges on the ability of the data to satisfy decomposability to a certain extent[23].

The concept of decomposability involves breaking down complex data into simpler components or features that models can learn and recombine at various levels of abstraction. This approach leverages the natural structure or patterns within the data. For instance, in image processing, lower layers of a model might capture basic features like edges and textures, while higher layers aggregate these features into more abstract representations such as shapes and objects.

Similarly, molecules exhibit inherent decomposability, which allows for hierarchical coding. The t-SMILES algorithm takes advantage of this by incorporating explicit hierarchical structure information into linear molecular representations using a tree-based approach. This hierarchical representation aligns with the natural decomposability of molecules, enabling more effective modeling and understanding of their complex structures.

With regard to the experimental outcomes on ChEMBL, TSIS and TSISR serve as illustrative examples due to their identical underlying models and symbols, differing only in the hierarchical structure of their sequences. This controlled setup helps eliminate potential interfering factors. Generally, TSIS consistently achieves higher FCD scores compared to TSISR.

Some researchers propose that the core of representation learning is the creation of multi-level feature representations with sufficient depth[33]. Additionally, others argue that the ability to capture hierarchical information in sequential data is crucial for developing intelligent systems capable of understanding and processing language[34]. In light of these considerations, our approach may provide valuable insights that can advance the field of molecular string representations.

**Compositional Generalization**

Human language learning benefits from a favorable type of combinatorial explosion: if a person knows the meaning of "to run" and that of "slowly", they can immediately understand what it means "to run slowly", even if they have never encountered this phrase before[35]. Compositionality is the classic idea that new representations can be constructed through the combination of primitive elements[36], which enables natural languages to construct complex semantic meanings from the combinations of simpler semantic elements[37][38]. Even with just a few examples, people can learn remarkably rich conceptual models[36]. Despite a multitude of empirical studies, little consensus exists on whether neural networks are able to generalise compositionally[21][36][39][40][41][42][43].

In the t-SMILES framework, if it is assumed that deep generative models lack the capacity to perform compositional generalization, there is still the possibility of achieving this when converting t-SMILES to SMILES. It is evident that when reconstruction is performed directly on training data, compositional generalization is the underlying process. It is reasonable to anticipate that the implementation of an advanced reconstruction algorithm, such as the goal-directed method[11], would result in superior outcomes. The second level of compositional generalization for t-SMILES is based on the capacity of neural networks. If aassuming that deep neural networks possess at least a limited notion of compositionality[36], it is reasonable to infer that the t-SMLES system would have compositional generalization ability from inner-fragments and inter-fragments.

In this scenario, the fragment-based t-SMILES algorithm achieves higer novelty score when getting same FCD score if model is well trained, which seems to be a more competitive linear descriptor than SMILES. An intriguing area of future research would be the investigation of the degree of compositional generalization achieved by these algorithms.

**Scaling Law**

As prior work[44] suggested, one fundamental characteristic of LLMs is the scaling law, which describes how the performance of LLMs improves as they scale in terms of model size, training data, and computational resources. This power-law relationship suggests that larger models trained on more data tend to capture more complex patterns and generalize better to new tasks[45][46].

As illustrated in Tables 2, models that undergo training on a greater volume of data tend to exhibit both higher novelty and FCD scores for all tested LMRs. This indicates that more complex patterns have been captured when training more. Similarly, the experiments on DFMs demonstrate that larger SMILES-based model DFM_SMILES[L12] achieves higher Novelty-FCD scores than smaller SMILES-based model DFM_SMILES[L8].

However, larger SMILES based diffusion model obtains lower Novelty-FCD scores than smaller t-SMILES model. This implies that complex patterns may be more efficiently learned from t-SMILES code than SMILES code by this diffusion model.

**Pre-trained then Fine-tuned**

Recent years have featured a trend towards pre-trained language representations in NLP systems, applied in increasingly flexible and task-agnostic ways for downstream transfer. In this paradigm, models are designed to be large to absorb information during pre-training, and are then fine-tuned on very narrow task distributions. There is evidence suggestting that the generalization achieved under this paradigm can be poor because the model is overly specific to the training distribution and does not generalize well outside it [47][48]. Thus, the performance of fine-tuned models on specific benchmarks, even when it is nominally at human-level, may exaggerate actual performance on the underlying task[49][50]. So the research[51] findings indicate a preference for few-shot learning over the necessity of explicit fine-tuning.

The few-shot model from a previous study[11] on a low-resource dataset demonstrated that even without data augmentation or pre-training, a standalone TSSA-based model outperformed a pre-trained and fine-tuned model using SMILES. Additionally, the results from the RL model in this study show that the generated target molecules are located outside the distribution of the pre-training dataset. This outcome highlights the unique logic of the t-SMILES framework, which enables t-SMILES encoding to exhibit a high degree of compositional generalization.

Although the pre-trained then fine-tuned paradigm is a classical approach in chemistry and materials science applications—where experimental data is often limited[52], it raises a question: can meaningful results truly be obtained with just tens to hundreds of data points? This presents an intriguing avenue for future research.

**Comparation of different LMRs**

**SMILES vs t-SMILES**

The t-SMILES framework has the ability to unify classical SMILES as TS_Vanilla, making t-SMILES a superset of SMILES[11]. Although SMILES is still used to represent molecular substructure, t-SMILES cuts molecules into small pieces and abandons the algorithm of depth-first traversal of molecular trees, which fundamentally reduces the depth of nesting and the proportion of characters that must appear in pairs.

Furthermore, the t-SMILES algorithm organizes fragments hierarchically, simplifying the overall problem-solving process. Experiments and analyses have shown that this hierarchical structure not only accelerates learning but also provides a more detailed and comprehensive representation of the molecular information.

**SELFIES**

A comparative analysis of models based on SELFIES, SMILES, and t-SMILES indicates that SELFIES-based models generally exhibit lower performance.

The relatively low scores observed in SELFIES models can be attributed to their tendency to generate molecules with fewer aromatic rings, which presents a challenge in learning the semantics. In other words, SELFIES models may produce more novel molecules if the similarity to the training data is disregarded. This observation is also consistent with the findings of other researchers, such as those reported in ref[53]. With regard to the limitations of SELFIES, a recent study suggested that learning the complete removal of valency constraints, which the researcher designated "unconstrained SELFIES," has been demonstrated to enhance performance outcomes[54].

**SAFE vs t-SMILES**

When comparing the algorithms of SAFE and t-SMILES, three perspectives become apparent:

1. Paired numbers can be divided into two categories.

The first category is the numbers that denote ring breaks in the classical SMILES grammar, and the second category is the numbers that denote cut points. The SAFE algorithm combines these two categories into one, which creates a semantic challenge. Nevertheless, the model must still learn and classify them.

2. Matching and dependency of paired numbers

The dependency length of the first type of number in TSID is limited within the fragment. Additionally, the dependency length of the second type of number [n*] can span across fragments. It is important to note that SAFE does not categorize the numbers, Therefore, the matching lengths and dependency lengths of the numbers can be interpreted as whole strings. Because SAFE is longer than SMILES, resulting in longer grammar dependencies in SAFE than in SMILES. It is worth noting that in both SAFE and TSID, the matching range of parentheses is limited to one fragment. So, in summary, SAFE does not address the long dependencies in SMILES syntax and may even increase the difficulty. For Example:

1) c1**3**cc(F)c(O)c(C2=CCc**3**c-2c(C)nn2C)c1.CO**3** ： Error: three symbols '3' that should be in pairs.
2) 'N15CCC**3**CCC1)CCCO2.Cc1cccc4c15.c16ccc7o1.C**3**6=O.C7(F)F.N**3**4': Error: unpaired symbols: ')', '2'and '3'.
3) 'c1**%18**c[nH]c2ccccc12.c1**7**ncc(Cl)c**%10**n1.N16CCCC9C1.C56=O.CC**7**.N5**7**.N**8**9': Error: unpaired symbols: '%18', '%10', '7', and '8'.

3. Molecule tree and sorting fragments based on size

Both foundational patterns can be learned by deep learning models. Sorting fragments based on size in SAFE may increase long-term dependencies. For example:

SMILES:

CCCCC1C(=O)N(C(C2SCCCS2)C2OC(C)(C)OC2C2COC(C)(C)O2)C1C(C)=Cc1ccccc1

The length of the string is 67.

SAFE：

C1**6**OC(C)(C)OC1C1COC(C)(C)O1.C17C(=O)N4C1C=3C.C=3c1ccccc1.C15SCCCS1.CCCC7.C45**6**

The length of the string is 77, which means that the dependency of the symbol '6' becomes longer.

**t-SMILES family**

The t-SMILES family has been expanded to include four major categories of codes: TSSA, TSDY, TSID, and TSIS. The TSIS code can be easily expanded to TSISD, TSISO, and TSISR through the application of straightforward rules.These seven types of code will be discussed from two perspectives: the complexity of the algorithms used and the determinacy of the information conveyed.

In consideration of the degree of complexity of code algorithms, TSSA is the most complex, employing shared atoms to link two parts. This can be evidenced by the observation that all TSSA-based models receive the lowest FCD scores across the entire t-SMILES family. TSDY and TSID are the next most complex, followed by all TSIS, including its variants, which are the simplest in comparison to the other three types of code.

1) TSSA vs TSDY/TSID: The distinction between the two lies in the inclusion of additional uncertainty information within the TSSA code. To illustrate, if the link point is designated as "C-N," then all atom C in one sug-fragment and all atom N in another are potential candidates for the linking point, thereby enhancing the possibility of producing novel molecules. However, parsing TSSA code is relatively more challenging than TSDY and TSID (lower FCD scores in experiment results). This indicates that either a more comprehensive model or longer training is necessary.

2) TSDY vs TSID: From the visualization in the diffusion model, it is apparent that the curve of TSID is covered by TSDY. A comparison of the metrics reveals that the TSDY model exhibits higher novelty and lower FCD scores than the TSID model. The distinction between TSDY and TSID can be attributed to the inclusion of more definitive information within TSID, derived from the linking token "[n*]". This observation prompts the thought that whether this difference is the primary reason that TSID is more efficiently learned than TSDY with a higher FCD score, but TSDY produces a greater number of novelty molecules.

3) TSIS vs TSID: The distinction between TSIS and TSID hinges on the FBT structure, which introduces a greater degree of definitive information within TSID. As illustrated in the sntax section, FBT is not an obstacle but may, in fact, be a beneficial element. However, there is no obvious statistical difference between the two that can be obversed from the experimental results on ChEMBL.

4) TSIS vs TSISR: The distinction between TSIS and TSISR is that TSIS uses the BFS algorithm to parse AMT, whereas TSISR randomly sorts sub-fragments. However, the TSISR models consistently exhibit the lowest FCD scores across all CLMS on ChEMBL. This finding suggests that incorporating hidden hierarchical and deterministic information can improve the effectiveness of linear molecular representations. By capturing underlying structural relationships, models can better represent and process molecular data, leading to enhanced performance in various tasks.

In conclusion, for relatively larger molecules, such as those in ChEMBL, hierarchical information is essential for effective data analysis and interpretation. Specifically, TSIS may achieve performance comparable to TSID. However, among the TSIS variants, TSISR is more difficult to parse than others due to its randomly ordered structure.

In light of the findings of the previous study[11] and the present one, it can be concluded that TSSA represents the optimal choice for goal-directed tasks. This is due to its superior generalization capacity, which enables it to circumvent "striking similarity" to the training dataset while achieving higher novelty with a reasonable degree of similarity. TSDY emerges as a balanced approach, suitable for both goal-oriented and distribution reproduction tasks. For distribution reproduction experiments, TSIS and TSID are recommended, as they demonstrate strong alignment with the physicochemical properties of the training data.

**Correction process in t-SMILES algorithm**

Despite the shorter length of the sub-fragment in t-SMILES compared to SMILES on average, as well as the reduced long-decency, the t-SMILES algorithm uses specific correction processes to guarantee the solution is robust.

1. If a sub-fragment is unable to be parsed as a valid fragment-mol, a fragment from the dictionary that is the most similar to the invalid sub-fragment is selected as a replacement.
2. If the link points [n*] in the TSID and TSIS code are unable to be correctly matched, the TSDY style is used as a corrective measure.
3. If all nodes of FBT or AMT could not be assembled as a single molecule due to the assembled graph conflicting with chemical knowledge, some smaller sub-fragments would be discarded, though this scenario is an infrequent occurrence when the model is adequately trained.

## Conclusion

In this study, we introduced a novel algorithm, TSIS, along with its variants—TSISD, TSISO, and TSISR—to the t-SMILES family. As a result, the t-SMILES framework now encompasses four fundamental encodings:TSSA, TSDY, TSID, and TSIS. A series of controlled experiments conducted using GPT, DFM, and RL demonstrate that the hierarchical structure inherent to the t-SMILES framework markedly enhances the convergence and generalization abilities of linear molecular representations.

Specifically, the hierarchical structure is not only relatively straightforward to parse but also empowers deep generative models to explore a broader chemical space, achieve higher novelty scores compared to SMILES, fit training data more accurately, and converge more rapidly in goal-oriented molecular generation tasks.

In contrast, SAFE faces some considerable challenges due to its long-term dependency issues, despite being a fragment-based algorithm. Furthermore, SELFIES based models consistently generate a higher proportion of molecules with fewer aromatic rings and larger rings, representing a notable divergence from the training data.

This is the second study on t-SMILES framework. Future research could explore additional areas such as interpretability, properties, retrosynthesis and reaction prediction, materials science, and other related fields. Additionally, the core algorithm of t-SMILES could be implemented in a faster programming language, such as C++, to enhance performance.

**Conflicts of Interest**

The authors declare that they have no competing interests.

**Author Contributions**

Juanni Wu and Ruqin Yu designed the study and manuscript. As the main designer of the project, Juanni Wu conceived the project, constructed the algorithms and Python script, performed the experiments, informatics analyses, and wrote the draft manuscript. Tong Wang, Lijuan Tang and Hailong Wu participated in the discussion and funding acquisition. All authors contributed to manuscript editing, revising and have approved the final version of the manuscript.

**Acknowledgements**

This research was funded by the National Natural Science Foundation of China.

**Data availability**

The datasets used in this study are publicly available. The processed data used in this study can be found at: https://github.com/juanniwu/t-SMILES/

**Code availability**

Code, training and generation scripts for this work can be found at: https://github.com/juanniwu/t-SMILES/